\documentclass[conf]{IEEEtran}
\usepackage{amsmath} 
\usepackage{gensymb}
\usepackage{cite}
\usepackage{balance}
\usepackage{float}
\usepackage{romannum}
\usepackage{graphicx}
\usepackage{comment}
\usepackage{subcaption}
\usepackage{placeins}
\usepackage{algorithm}
\usepackage{algorithmic}
\usepackage{multirow}
\usepackage{amsmath}
\usepackage{amssymb}
\usepackage{amsxtra}
\usepackage{multirow}
\usepackage{mathtools}
\usepackage{color}
\usepackage{bbm}
\title{Empowering Urban Traffic Management: Elevated 3D LiDAR for Data Collection and Advanced Object Detection Analysis}
\author{\IEEEauthorblockN{\large Nawfal Guefrachi$^{1}$, Hakim Ghazzai$^{2}$, and Ahmad Alsharoa$^{1}$}\\
\IEEEauthorblockA{{\small $^{1}$Missouri University of Science and Technology (MST), Rolla, Missouri, USA\\
$^{2}$CEMSE Division, King Abdullah University of Science and Technology (KAUST), Thuwal, Saudi Arabia}}\vspace{-1.15cm}
}
\begin{document}
\vspace{-0.5cm}
\maketitle
\thispagestyle{empty}
\begin{abstract}
\boldmath
The 3D object detection capabilities in urban environments have been enormously improved by recent developments in Light Detection and Range (LiDAR) technology. This paper presents a novel framework that transforms the detection and analysis of 3D objects in traffic scenarios by utilizing the power of elevated LiDAR sensors. We are presenting our methodology's remarkable capacity to collect complex 3D point cloud data, which allows us to accurately and in detail capture the dynamics of urban traffic. Due to the limitation in obtaining real-world traffic datasets, we utilize the simulator to generate 3D point cloud for specific scenarios. To support our experimental analysis, we firstly simulate various 3D point cloud traffic-related objects. Then, we use this dataset as a basis for training and evaluating our 3D object detection models, in identifying and monitoring both vehicles and pedestrians in simulated urban traffic environments. Next, we fine tune the Point Voxel-Region-based Convolutional Neural Network (PV-RCNN) architecture, making it more suited to handle and understand the massive volumes of point cloud data generated by our urban traffic simulations. Our results show the effectiveness of the proposed solution in accurately detecting objects in traffic scenes and highlight the role of LiDAR in improving urban safety and advancing intelligent transportation systems.
\end{abstract}

\vspace{-0.15cm}
\vspace{-0.4cm}
\section{Introduction}
Real-time and precise object detection plays a critical role in traffic surveillance by providing accurate object detection in a Three-Dimensional (3D) environment. Public safety can be improved by monitoring the behavior of vehicles and pedestrians through the integration of Internet-of-Things (IoT) sensors with existing infrastructure~\cite{ahmed2022smart}. Traditional surveillance methods mostly relied on standard or depth cameras which are often inadequate in complex or congested areas due to illumination requirements and privacy concerns ~\cite{nguyen2016human, hung2020faster}.

There are several issues with the current state of 3D object detection systems in traffic surveillance. Standard camera-based surveillance systems are significantly less effective in complicated or densely inhabited locations due to inadequate lighting and privacy concerns~\cite{7846643}. On the other hand, depth cameras can help in pedestrian detection by using depth information to partition head regions in crowded areas, however, they struggle to tell apart overlapping objects in high-density environments~\cite{liciotti2017people}.

Light Detection and Ranging (LiDAR) technology has become increasingly important for real-time object detection and classification, offering the capability to create 3D maps of the surrounding environments through laser reflections~\cite{rinchi2023LiDAR}. Its precision in tracking movements of objects like pedestrians and vehicles supports privacy-conscious decisions in various applications, from parking security to traffic management~\cite{yang2024LiDAR}. Further, LiDAR sensors can operate effectively under any lighting conditions while safeguarding Personal Identifiable Information (PII)~\cite{guefrachi2024leveraging}. Several works proposed using LiDAR's to enhance perception capabilities through 3D object detection ~\cite{10181371}. In~\cite{8569311}, the authors presented a LiDAR-based framework that employs cell encoding and Convolutional Neural Networks (CNNs) techniques for detecting object locations and orientations. Furthermore, the authors in~\cite{9096075} proposed a CNN and supervised learning LiDAR-based approaches for 3D object detection that involve data syncing, ground detection, region definition, clustering, and occupancy grid mapping.

The development of autonomous driving technology has been largely dependent on the usage of real-world datasets, such as those from Waymo or Kitti~\cite{liao2022kitti, sun2020scalability}. However, these datasets often do not capture the entire range of real-world situations, especially in complex metropolitan settings where pedestrian behavior is important. Moreover, these datasets and associated models are mainly focusing on autonomous navigation purposes including autonomous driving and collision avoidance. These shortcomings highlight the need for more comprehensive datasets to advance the robustness of autonomous driving technology in varied real-world conditions. 

In this paper, we address the shortcomings of conventional data collection techniques by investigating the complexities of pedestrians and vehicles detection in urban environments using a range of simulation platforms and data generation techniques. In particular, we use simulators to create lifelike urban scenes and animate dynamic components such as pedestrians and vehicles. By using this method, we produce datasets that are customized for precise  objects detection in 3D point clouds applications, which makes it easier to train and assess deep learning models for 3D point clouds processing such as Point Voxel-Region-based Convolutional Neural Network (PV-RCNN) and Sparsely Embedded Convolutional Detection (SECOND). Our results show that the use of custom data labeling techniques and simulation tools can improve the precision of vehicle and pedestrian detection, which can impact on the advancement of urban surveillance systems. This paper provides a concise overview about the effectiveness of employing static/elevated LiDAR sensors for traffic monitoring and the efficiency of use of simulated data collection techniques to build a custom dataset to train deep learning models. 

\begin{figure}[t!]
\centering
 \includegraphics[width=2.1in]{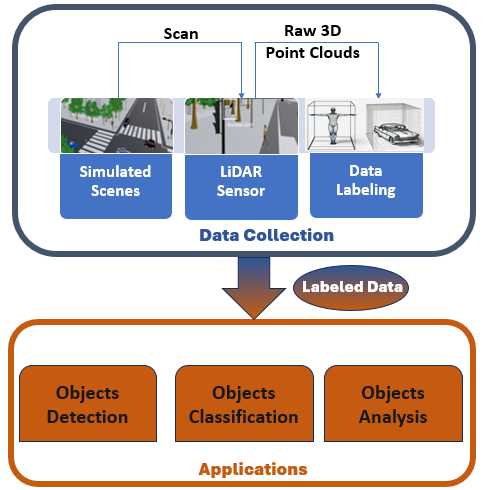}
\caption{General workflow of object monitoring.}
\label{fig1}
\vspace{-7mm}
\end{figure}
\section{LiDAR-enabled Tasks and Data Scarcity}
\vspace{-1mm}
LiDAR technology has emerged as a pivotal tool for urban monitoring, significantly enhancing pedestrian and vehicle detection as well as environmental analysis. Its high-resolution capabilities enable the precise identification and localization of objects within complex urban landscapes, an important feature for real-time security monitoring and identifying potential security violations. Furthermore, it aids urban planners by providing insights into creating safer, more pedestrian-oriented urban environments, thus contributing to superior urban design. Among the primary tasks of LiDAR in urban traffic management, like it is illustrated in Figure \ref{fig1}, we can list:

$\bullet$ \textbf{Object detection}: Particularly when focused on identifying vehicles and pedestrians, object detection is pivotal for applications requiring real-time situational awareness and decision-making, such as autonomous driving and urban surveillance. By utilizing bounding boxes to locate and classify each object as either a vehicle or a pedestrian within a scene, this technology sets the stage for advanced subsequent applications such us activity classification, gait analysis and trajectory prediction.

$\bullet$ \textbf{Activity classification}: It extends beyond identifying objects like pedestrians and vehicles; it categorizes their behaviors, enhancing urban safety and efficiency. For pedestrians, it distinguishes actions such as walking or running, essential for smart city safety initiatives like adaptive lighting. For vehicles, it identifies driving profiles such as overspeeding, which is pivotal for intelligent traffic management and accident prevention. This capability is key in smart cities because it allows for accurate understanding of dynamic behaviors and the creation of safer and more adaptable urban environments.

$\bullet$ \textbf{Gait analysis}: Which is specifically customized for pedestrians and uses LiDAR technology, provides valuable insights into individual movement patterns, thereby improving traffic management and transportation safety. This technology detects abnormal pedestrian behavior patterns, such as sudden deviations or unexpected stops, by analyzing the unique characteristics of a pedestrian's gait, which is critical for alerting and reducing potential dangers within urban transit systems. Although gait analysis has applications in health and security by identifying mobility issues and improving identification procedures, its role in traffic management is essential. It is a non-intrusive tool for ensuring pedestrian safety, helping to develop smarter, more responsive transportation infrastructures that can anticipate and prevent incidents before they happen.

$\bullet$ \textbf{Trajectory Prediction}: It is an important component of smart navigation systems, considers the orientation of vehicles and pedestrians in addition to their location. This comprehensive approach allows for accurate forecasting of future movements, which is important for avoiding incidents in dynamic environments. AI-powered systems, such as jetbots, can effectively anticipate potential collisions and change routes in real time because they understand both the direction and speed at which an object moves. Integrating orientation data into trajectory estimation improves prediction precision, allowing for safe and efficient navigation through complex urban landscapes. This capability is essential for deploying autonomous systems that can interact with human-centric urban spaces, ensuring both safety and smooth mobility.

The deployment of fixed LiDAR infrastructure for traffic management presents significant challenges, particularly in developing deep learning models that can adapt to a variety of settings and scenarios. A major impediment is a lack of comprehensive datasets required for accurate identification and localization of road users. Consistent data collection is challenging in urban environments due to their dynamic nature, demonstrated by ongoing development, varying pedestrian flows, and changing traffic patterns. These obstacles drive advanced research into pedestrian and driver behavior and the development of traffic monitoring, smart mobility, and urban planning methodologies. Furthermore, this scarcity reduces the algorithms' ability to adapt to complex behaviors and situations, emphasizing the importance of addressing data collection and environmental variability issues. Despite LiDAR technology's promise to revolutionize urban monitoring, overcoming these data-related challenges is critical to realizing its full potential.

\vspace{-4mm}
\section{Data Collection Strategies}
Point cloud data collection can be conducted in various ways: by gathering it directly from real-world experiments, utilizing simulators, or employing a combination of both approaches. Collecting data from the real world is often challenging and time-consuming, as it involves complex setups, requires approved authorizations, and is subject to unpredictable factors. On the contrary, simulators and experimental setups offer a more controlled environment, allowing researchers to easily create different environment conditions and generate data that may be difficult or impossible to obtain in the real world conditions.

\vspace{-4mm}
\subsection{Experimental Approach}
\begin{figure}[t!]
\centering
 \includegraphics[width=2in]{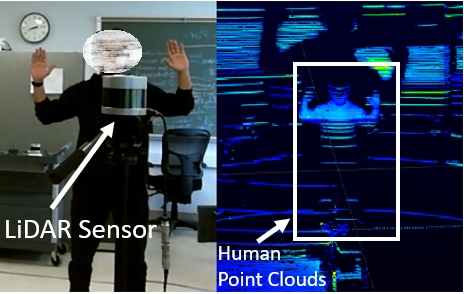}
\caption{Experimental setup for pedestrian detection, conducted in the OIWS lab at MST.} 
\label{fig5}
\end{figure}

Experimentation is vital for the development of autonomous systems, allowing for the assessment of designs and technologies in real-world scenarios. A small-scale LiDAR scanning system with a tripod-mounted sensor aimed at a designated region and a workstation displaying a 3D model are shown in Figure ~\ref{fig5}. The experiment closely monitors the speed, trajectory, and interactions of pedestrians. The result is a comprehensive dataset with temporal and spatial insights that are important for improving the safety of autonomous systems and modeling pedestrian behavior. This extensive dataset will be useful for researchers in examining movement patterns of pedestrians and performing gait analysis.

\begin{figure}[t!]
\centering
 \includegraphics[width=2in]{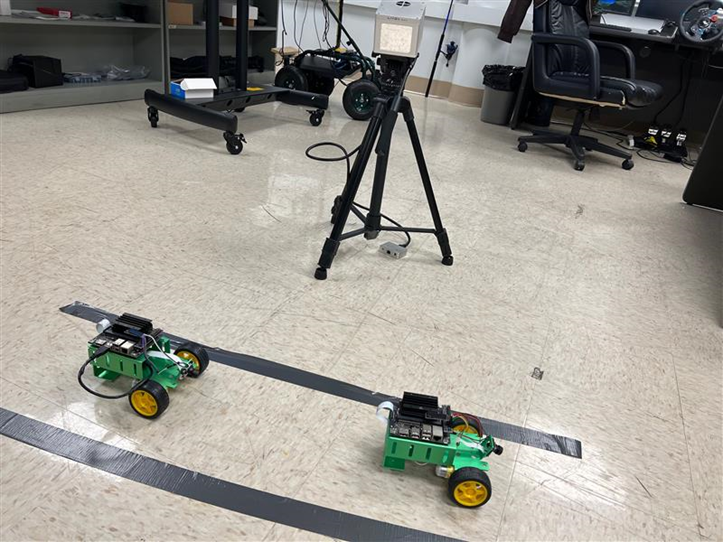}
\caption{Experimental setup for vehicle detection using Jetbots, conducted in the OIWS lab at MST.}
\label{fig6}
\vspace{-4mm}
\end{figure}

Figure \ref{fig6} illustrates an experimental setup utilizing SeedStudio Jetbot Smart Car Kits to simulate vehicle movements in a controlled setting. These kits, which include advanced sensors and modules, simulate traffic scenarios and generate data for unmanned vehicle and traffic pattern research. Researchers can tailor different conditions for data collection to specific real-world applications using programmable tasks and precise speed control. The jetbots, which are monitored by static LiDAR sensors, allow for a detailed traffic analysis, including acceleration, steering patterns, collision prevention, and the detection of abnormal driving behaviors. This makes it easier to create labeled datasets in controlled environments, which helps to develop deep learning models for real-world applications.


\vspace{-5mm}
\vspace{-1mm}
\subsection{Simulator Approach}
\vspace{-1mm}
Some platforms work well to simulate traffic situations and enable fast and effective data collection, but struggle to deal with complex urban environments and detailed pedestrian behavior to capture diverse and unpredictable scenarios. In the following, we present the simulators that can be used to implement the aforementioned tasks, integrate LiDAR sensors, and enable 3D point cloud data collection.

\subsubsection{Webots}
Webots, while useful in optimizing LiDAR settings for robotics and autonomous vehicle simulations, has limitations, most notably in the realism of its pedestrian and vehicle models. Despite these limitations, it is still useful for data collection and preliminary testing, with focused LiDAR adjustment increasing simulation precision for traffic management and autonomous driving.

\subsubsection{Gazebo}
For scene and sensor simulations, roboticists highly regard Gazebo, a simulation tool. It excels at correctly simulating complex robotic systems in particular. Still, it is difficult to model pedestrian behavior, which is important for applications like traffic control and self-driving car navigation.


\subsubsection{Carla}
CARLA, an open-source simulator designed for autonomous driving research, provides high-fidelity environments, diverse weather conditions, and lighting variations, making it an ideal tool for testing and validating autonomous vehicle systems across a variety of scenarios. It includes pedestrian models that improve the realism of urban simulations. Despite its vehicle simulation capabilities, CARLA's pedestrian behavior models fall short of accurately representing the full complexity of pedestrian actions, particularly abnormal activities.

\subsubsection{Blender for Data Generation}
In this work, we use Blender, an all-in-one 3D graphics software, to create scenes and collect data. Blender is a suitable option for creating detailed and realistic 3D images due to its extensive modeling, texturing, and animation toolkit. Its strong simulation capabilities enable artists to produce renders of exceptional quality and its versatility makes it appropriate for a broad range of applications, including 3D animation and real-world simulation.

Using add-ons like BLAINDER ~\cite{reitmann2021blainder}, we improve depth-sensing simulations for AI research, making it easier to create custom datasets for robotics and computer vision. Blender's node-based architecture enables complex effects and animations. By simulating the interaction of light and material, BLAINDER enables the generation of accurate depth and semantic data, which is critical for object detection and navigation applications. This approach enables customization and the creation of large datasets.

\vspace{-3mm}
\subsection{Hybrid Approach: Adjustable Simulator for Real-World Scenarios}
\begin{figure}[t]
  \centering
  \includegraphics[width=2in]{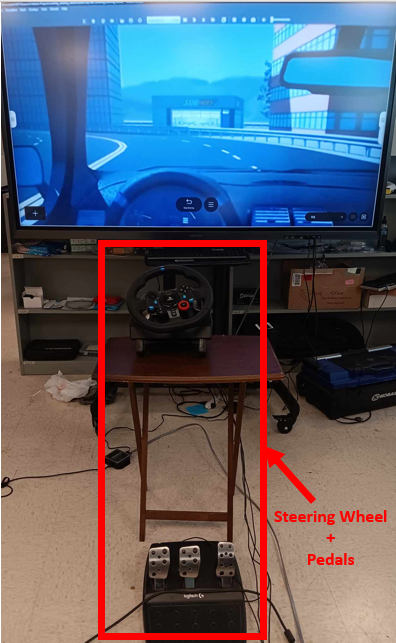}
  \caption{This work, conducted in the OIWS lab at MST, involved collecting LiDAR data through a steering wheel in a simulated environment.}
  \vspace{-1mm} 
  \label{hybrid}
\end{figure}

The hybrid approach combines simulations and real-world elements to improve research accuracy and relevance, as shown in Figure~\ref{hybrid}. This technique combines LiDAR-equipped virtual simulations with a physical steering wheel setup to collect data on vehicle movements in urban settings. The addition of a Driving Force G29 Gaming Racing Wheel and responsive pedals enhances realism, allowing for more detailed scenario analysis. This setup simulates both normal and unusual driving behaviors, such as irregular movements and sudden lane changes, as well as accurate vehicle control. Such detailed control enables in-depth vehicle dynamics and behavior analysis, yielding important insights into autonomous vehicle safety and urban navigation.
\vspace{-4mm}
\section{Data Generation}
In our urban environment simulation work, we used Blender to position scene objects and animate elements of interest such as pedestrians and vehicles, adding realism to our simulations. This method enabled us to create detailed and dynamic urban landscapes, which is essential to improve the accuracy and efficacy of our 3D object detection models.
\vspace{-5mm}
\subsection{Urban Environment Simulation and Scenarios}
\vspace{-1mm}
\subsubsection{Objects Animations}
We started by arranging the scene's objects and then animated the pedestrians and vehicles. We used rigging for pedestrian animation, which involves adding skeletons to 3D models to script movements and allow realistic poses by manipulating bones for body parts such as the spine and limbs. We used Inverse Kinematics (IK) to achieve more natural motion and streamlined the animation with keyframe adjustments for actions like walking and running. Vehicle animation, on the other hand, was more straightforward, requiring no rigging and instead relying on direct manipulation and keyframe animation to simulate realistic movement, ensuring seamless integration with the scene.
\begin{figure}[t]
  \centering
  \includegraphics[width=3in]{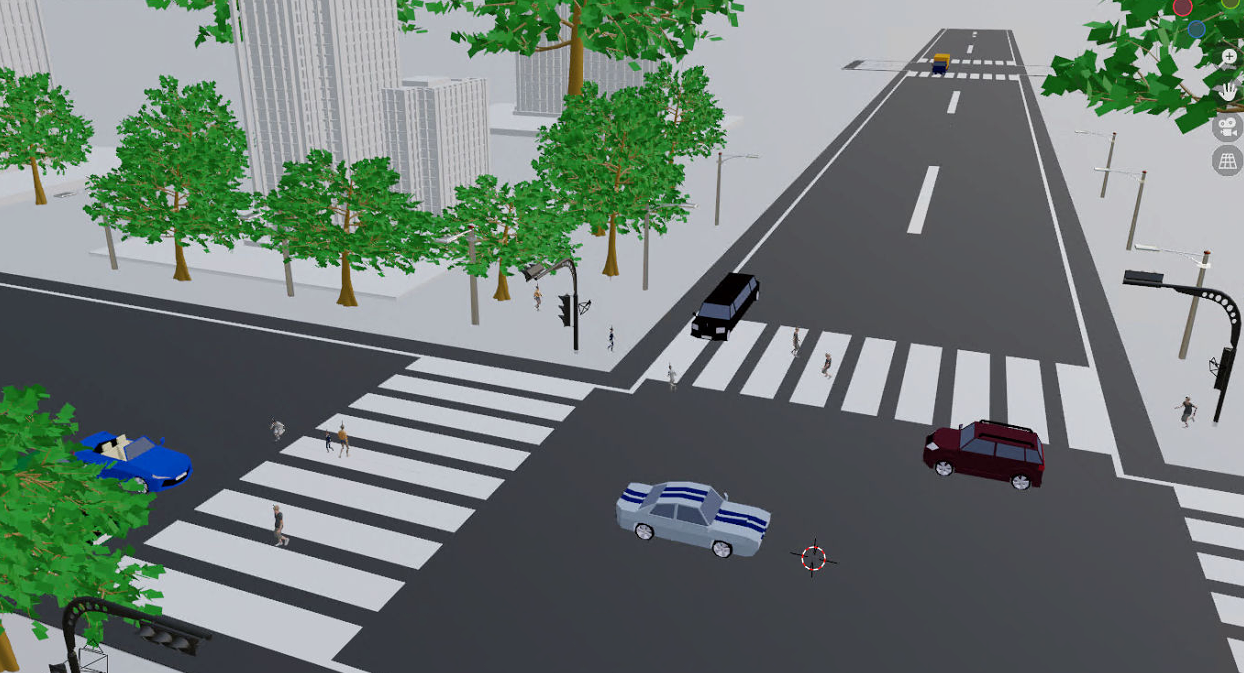}
  \caption{Screenshot of a simulated scene from Blender.}
  \vspace{-3mm} 
  \label{Scene}
\end{figure}

\subsubsection{Scenarios Specifications}
Our scenarios primarily focus on traffic movement. We simulate the environments with pedestrians and vehicles moving with certain speed, as shown in Figure \ref{Scene}. We animate pedestrians and vehicle characters using keyframes to specify their motion properties, such as starting and ending positions, speed, and trajectories. This will enable the Blender simulator to mimic realistic animations. A script facilitated parameter adjustments, particularly for pedestrian walking and running actions, integrating them into traffic to mimic real situations. Each scene showcases various vehicle types, such as SUVs, buses, and trucks moving in different directions and timings. We also included diverse pedestrian models of various genders and sizes, as well as three to four children who demonstrated behaviors such as running and walking at different speeds.
\vspace{-5mm}
\subsection{Data Collection}

\subsubsection{Collection of Raw Point Cloud Data}
Compiling real-world datasets for monitoring pedestrians and vehicles is challenging, especially when safety issues arise and dangerous situations such as falls are mimicked. We use Blender to recreate the required scenarios in order to avoid these issues. Blender is chosen because of its advanced 3D point cloud manipulation capabilities, which include the capacity to replicate LiDAR and animate pedestrians \cite{reitmann2021blainder}. With the ability to reflect beam intensities in different materials, this feature allows for accurate modeling of LiDAR data. To obtain coherent 3D coordinates and reflectivity of objects in traffic scenarios, we deploy several LiDAR sensors at strategic heights and angles. To augment our dataset for more advanced analysis, we use Blender to simulate vehicles and various human models engaged in different activities.
\vspace{1mm}
\subsubsection{Data Labeling}
To effectively process raw data, a thorough annotation of each data point is necessary. It is necessary to generate 3D bounding boxes that follow a predetermined format in order to accurately capture the locations of objects in different settings. This format includes class label identity, positions, and dimensions. This consistent and comprehensive foundation for data labeling offered by this method is essential for machine learning models to correctly detect and identify objects. By following a standardized format, it becomes easier to train models that can reliably recognize and differentiate between objects in various scenarios, leading to improved accuracy in object detection tasks.

\begin{figure}[t]
  \centering
  \includegraphics[width=2in]{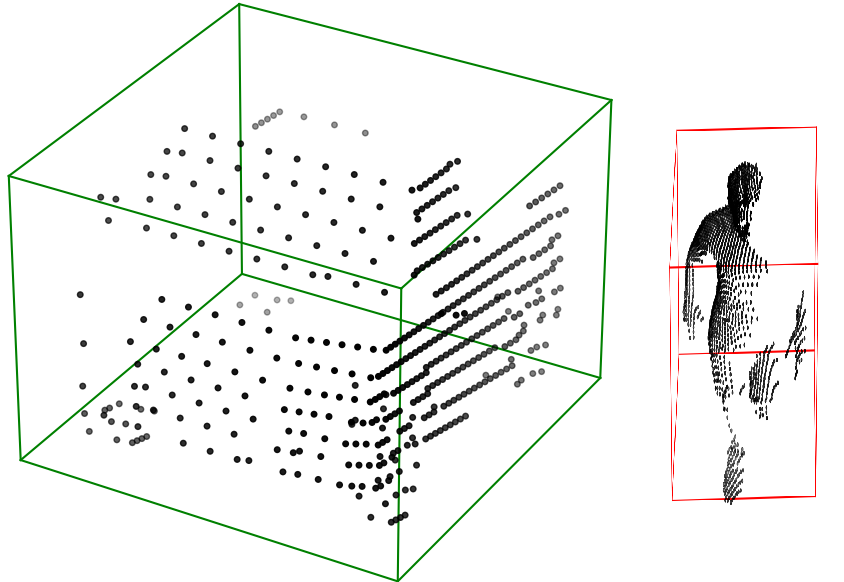}
  \caption{Dataset visualization: pedestrian and vehicle point clouds.}
  \label{vis1}
  \vspace{-5mm}
\end{figure}

\vspace{-4mm}
\section{Case of Study: Detection of pedestrians and vehicles using Blender Simulator}
\subsection{Dataset Specifications}
\vspace{-1mm}
Using Blender, we create 17 different scenes featuring a variety of vehicles, 10-20 pedestrians of different genders and body types, and 3-4 children, all engaged in a range of activities. These scenes, consisting of between 550 and 2500 frames, are designed to imitate real-world scenarios. Each scene is equipped with two or four LiDAR sensors mounted on traffic lights, which collect more than 350,000 3D point clouds. Our extensive data collection is well-suited for an accurate object detection application, as demonstrated by our 3D point cloud visualizations of vehicles and pedestrians (Figure \ref{vis1}). This comprehensive approach ensures that our simulations closely mirror actual urban environments, providing robust data for analysis and model training. 
\vspace{-4mm}
\subsection{Deep Learning Architectures}
\vspace{-1mm}
By analyzing raw LiDAR data, our Point Voxel-Region-based Convolutional Neural Network (PV-RCNN) architecture accurately identifies and localizes objects in point clouds, demonstrating high efficacy in 3D object detection. We improve the accuracy and flexibility of object location by adjusting network parameters, addressing data inconsistencies, and optimizing data set pre-processing. Accurate object representation is ensured by adjusting the voxel feature encoding and standardizing bounding box measurements. An important technique in our approach is Furthest Point Sampling (FPS), which selects 4096 keypoints to capture a large feature set for detailed environment analysis.

For comparison, a voxel-based architecture is utilized: Sparsely Embedded Convolutional Detection (SECOND)~\cite{yan2018second}. In order to extract significant features, it uses Voxel Feature Encoding (VFE) layers to transform 3D points input into voxels. The use of 3D convolutional layers comes next to recognize spatial patterns for objects identification.

To evaluate our object detection model's performance, we focus on a concise set of metrics: (1) Average Precision (AP) measures the model's true positive identification across thresholds. (2) Precision, emphasizing the true positive rate and the need to reduce false positives. (3) Recall, which measures the model's ability to detect all relevant objects. (4) The F1 score combines recall and precision to provide a comprehensive measure of the accuracy of the model in detecting the correct instances.
\vspace{-2mm}
\FloatBarrier
\begin{table}[htbp]
\centering
\caption{Detection results: PV-RCNN versus SECOND}
\label{tab:detection}
\begin{tabular}{|l|p{1.3cm}|p{1.3cm}|p{1.3cm}|p{1.3cm}|}
\hline
\textbf{Class} & \multicolumn{2}{c|}{\textbf{Pedestrians}} & \multicolumn{2}{c|}{\textbf{Vehicles}} \\
\hline
\textbf{Metric} & \textbf{PV-RCNN} & \textbf{SECOND} & \textbf{PV-RCNN} & \textbf{SECOND} \\
\hline
AP & \textbf{82.27\%} & 73.89\% & \textbf{85.36\%} & 76.33\% \\
\hline
Precision & \textbf{83.76\%} & 73.89\% & \textbf{85.36\%} & 76.12\% \\
\hline
Recall & \textbf{80.89\%} & 70.36\% & \textbf{83.72\%} & 74.58\% \\
\hline
F1-Score & \textbf{82.30\%} & 72.07\% & \textbf{84.52\%} & 75.32\% \\
\hline
\end{tabular}
\vspace{-0.8cm}
\end{table}
\subsection{Detection Results}
\vspace{-1mm}
When it comes to identifying vehicles and pedestrians, PV-RCNN outperforms SECOND in our comparison of the two 3D object detection models, as demonstrated in Table \ref{tab:detection}. Specifically, PV-RCNN achieves higher F1-Scores of 82.30\% and 84.52\% for pedestrian and vehicle detection, respectively, compared to SECOND's scores of 72.07\% and 75.32\%. These findings underscore the superior detection capabilities of PV-RCNN, attributable to its hybrid architecture in integrating point and voxel features within 3D point clouds, thereby enhancing its performance even in densely populated urban environments. The accuracy of the model and the high recall rates validate its robustness. The PV-RCNN model is unique in the domain of vehicle detection in urban settings, demonstrating a remarkable ability to identify a wide range of vehicles with robustness. Its ability to recognize pedestrians makes it an invaluable tool for urban surveillance and safety applications. The PV-RCNN model improves the ability to monitor and analyze urban traffic dynamics by combining detection precision with resilience to changing environmental conditions, which is important to advance urban safety measures.

\vspace{-4mm}
\section{Conclusion}
\vspace{-1mm}
In this paper, the goal was to improve 3D object detection in urban traffic using cutting-edge LiDAR technology and an improved PV-RCNN model based on a custom Blender dataset. This dataset was important because it simulated a variety of realistic urban scenarios involving vehicles and pedestrians, providing our model with extensive data that significantly improved detection precision. Our findings demonstrated the critical role of comprehensive datasets in the development of effective 3D detection systems and illustrated the ways in which LiDAR technology establishes the foundation for further advancements in pedestrian safety protocols.
\bibliographystyle{IEEEtran}\balance
\vspace{-3mm}
\bibliography{References}

\begin{thebibliography}{10}
\providecommand{\url}[1]{#1}
\csname url@samestyle\endcsname
\providecommand{\newblock}{\relax}
\providecommand{\bibinfo}[2]{#2}
\providecommand{\BIBentrySTDinterwordspacing}{\spaceskip=0pt\relax}
\providecommand{\BIBentryALTinterwordstretchfactor}{4}
\providecommand{\BIBentryALTinterwordspacing}{\spaceskip=\fontdimen2\font plus
\BIBentryALTinterwordstretchfactor\fontdimen3\font minus \fontdimen4\font\relax}
\providecommand{\BIBforeignlanguage}[2]{{%
\expandafter\ifx\csname l@#1\endcsname\relax
\typeout{** WARNING: IEEEtran.bst: No hyphenation pattern has been}%
\typeout{** loaded for the language `#1'. Using the pattern for}%
\typeout{** the default language instead.}%
\else
\language=\csname l@#1\endcsname
\fi
#2}}
\providecommand{\BIBdecl}{\relax}
\BIBdecl

\bibitem{ahmed2022smart}
I.~Ahmed, G.~Jeon, and A.~Chehri, ``{A Smart IoT Enabled End-to-End 3D Object Detection System for Autonomous Vehicles},'' \emph{{IEEE Transactions on Intelligent Transportation Systems}}, vol.~24, no.~11, pp. 13\,078 -- 13\,087, Oct. 2022.

\bibitem{nguyen2016human}
D.~T. Nguyen, W.~Li, and P.~O. Ogunbona, ``{Human Detection from Images and Videos: A survey},'' \emph{{Pattern Recognition}}, vol.~51, pp. 148--175, March. 2016.

\bibitem{hung2020faster}
G.~L. Hung, M.~S.~B. Sahimi, H.~Samma, T.~A. Almohamad, and B.~Lahasan, ``{Faster R-CNN Deep Learning Model for Pedestrian Detection from Drone Images},'' \emph{{SN Computer Science}}, vol.~1, pp. 1--9, April. 2020.

\bibitem{7846643}
L.~Hou, W.~Wan, K.~Han, R.~Muhammad, and M.~Yang, ``{Human Detection and Tracking over Camera Networks: A Review},'' in \emph{proc. of the 2016 International Conference on Audio, Language and Image Processing (ICALIP), Shanghai, China}, July. 2016, pp. 574--580.

\bibitem{liciotti2017people}
D.~Liciotti, M.~Paolanti, E.~Frontoni, and P.~Zingaretti, ``{People Detection and Tracking from an RGB-D Camera in Top-View Configuration: Review of Challenges and applications},'' in \emph{proc. of the ICIAP International Workshops, WBICV, SSPandBE, 3AS, RGBD, NIVAR, IWBAAS, and MADiMa, Catania, Italy}, Dec. 2017, pp. 207--218.

\bibitem{rinchi2023LiDAR}
O.~Rinchi, H.~Ghazzai, A.~Alsharoa, and Y.~Massoud, ``{LiDAR Technology for Human Activity Recognition: Outlooks and Challenges},'' \emph{IEEE Internet of Things Magazine}, vol.~6, no.~2, pp. 143--150, Feb. 2023.

\bibitem{yang2024LiDAR}
W.~Yang, D.~Li, W.~Xu, and Z.~Zhang, ``{A LiDAR-Based Parking Slots Detection System},'' \emph{{International Journal of Automotive Technology}}, vol.~11, no.~3, pp. 1--8, Feb. 2024.

\bibitem{guefrachi2024leveraging}
N.~Guefrachi, J.~Shi, H.~Ghazzai, and A.~Alsharoa, ``{Leveraging 3D LiDAR Sensors to Enable Enhanced Urban Safety and Public Health: Pedestrian Monitoring and Abnormal Activity Detection},'' in \emph{{proc. of the 46th Annual International Conference of IEEE Engineering in Medicine and Biology Society, Orlando, Florida, USA}}, July. 2024.

\bibitem{10181371}
B.~Cherif, H.~Ghazzai, A.~Alsharoa, H.~Besbes, and Y.~Massoud, ``{Aerial LiDAR-based 3D Object Detection and Tracking for Traffic Monitoring},'' in \emph{proc. of the 2023 IEEE International Symposium on Circuits and Systems (ISCAS), Monterey, USA}, July. 2023.

\bibitem{8569311}
J.~Beltrán, C.~Guindel, F.~M. Moreno, D.~Cruzado, F.~García, and A.~De~La~Escalera, ``{BirdNet: A 3D Object Detection Framework from LiDAR Information},'' in \emph{proc. of the 21st International Conference on Intelligent Transportation Systems (ITSC), Maui, HI, USA}, Nov. 2018.

\bibitem{9096075}
J.~N. Hayton, T.~Barros, C.~Premebida, M.~J. Coombes, and U.~J. Nunes, ``{CNN-Based Human Detection Using a 3D LiDAR onboard a UAV},'' in \emph{proc. of the 2020 IEEE International Conference on Autonomous Robot Systems and Competitions (ICARSC), Ponta Delgada, Portugal}, Apr. 2020, pp. 312--318.

\bibitem{liao2022kitti}
Y.~Liao, J.~Xie, and A.~Geiger, ``{Kitti-360: A Novel Dataset and Benchmarks for Urban Scene Understanding in 2D and 3D},'' \emph{{IEEE Transactions on Pattern Analysis and Machine Intelligence}}, vol.~45, no.~3, pp. 3292--3310, June. 2022.

\bibitem{sun2020scalability}
P.~Sun, H.~Kretzschmar, X.~Dotiwalla, A.~Chouard, V.~Patnaik, P.~Tsui, J.~Guo, Y.~Zhou, Y.~Chai, B.~Caine \emph{et~al.}, ``{Scalability in Perception for Autonomous Driving: Waymo Open Dataset},'' in \emph{{proc. of the IEEE/CVF conference on computer vision and pattern recognition, Washinton, USA}}, June. 2020.

\bibitem{reitmann2021blainder}
S.~Reitmann, L.~Neumann, and B.~Jung, ``{Blainder a Blender AI Add-on for Generation of Semantically Labeled Depth-Sensing Data},'' \emph{{Sensors}}, vol.~21, no.~6, p. 2144, March. 2021.

\bibitem{yan2018second}
Y.~Yan, Y.~Mao, and B.~Li, ``{Second: Sparsely embedded convolutional detection},'' \emph{{Sensors}}, vol.~18, no.~10, p. 3337, Oct. 2018.

\end{thebibliography}
\vspace{-5mm}
\end{document}